\documentclass[runningheads]{llncs}
\usepackage[T1]{fontenc}
\usepackage[table]{xcolor}
\usepackage{graphicx}
\usepackage{booktabs}
\usepackage{amsmath}
\usepackage{hyperref}
\usepackage{todonotes}
\usepackage{multirow}

\begin{document}
%
\title{nnU-Net Revisited: \\ A Call for Rigorous Validation \\ in 3D Medical Image Segmentation}
%
\titlerunning{nnU-Net Revisited}
%
\author{
Fabian Isensee$^{*}$\inst{1,3}, Tassilo Wald$^{*}$\inst{1,3,7}, Constantin Ulrich$^{*}$\inst{1,5,6}, Michael Baumgartner$^{*}$\inst{1,3,7}, Saikat Roy\inst{1}, Klaus Maier-Hein$^{\dag}$\inst{1,3,4,5,6,7}, Paul Jäger$^{\dag}$\inst{2,3}}

\authorrunning{Isensee \& Wald \& Ulrich \& Baumgartner et al.}

%
\institute{
Division of Medical Image Computing, German Cancer Research Center (DKFZ), Heidelberg, Germany  \and 
Interactive Machine Learning Group (IML), DKFZ, Heidelberg, Germany \and 
Helmholtz Imaging, DKFZ, Heidelberg, Germany. \and 
Pattern Analysis and Learning Group, Department of Radiation Oncology, Heidelberg University Hospital, Heidelberg, Germany \and 
National Center for Tumor Diseases (NCT) Heidelberg, Germany \\
\and Medical Faculty Heidelberg, University of Heidelberg, Heidelberg, Germany\\
\and Faculty of Mathematics and Computer Science, University of Heidelberg, Heidelberg, Germany\\
\email{f.isensee@dkfz-heidelberg.de}
}

\maketitle              

\begin{abstract}
The release of nnU-Net marked a paradigm shift in 3D medical image segmentation, demonstrating that a properly configured U-Net architecture could still achieve state-of-the-art results. Despite this, the pursuit of novel architectures, and the respective claims of superior performance over the U-Net baseline, continued. In this study, we demonstrate that many of these recent claims fail to hold up when scrutinized for common validation shortcomings, such as the use of inadequate baselines, insufficient datasets, and neglected computational resources. By meticulously avoiding these pitfalls, we conduct a thorough and comprehensive benchmarking of current segmentation methods including CNN-based, Transformer-based, and Mamba-based approaches. In contrast to current beliefs, we find that the recipe for state-of-the-art performance is 1) employing CNN-based U-Net models, including ResNet and ConvNeXt variants, 2) using the nnU-Net framework, and 3) scaling models to modern hardware resources. These results indicate an ongoing innovation bias towards novel architectures in the field and underscore the need for more stringent validation standards in the quest for scientific progress.

\end{abstract}
\def\thefootnote{*}\footnotetext{Equal contribution. Authors are permitted to list their name first in their CVs.}\def\thefootnote{\arabic{footnote}}
\def\thefootnote{\dag}\footnotetext{Equal supervision.}\def\thefootnote{\arabic{footnote}}
\section{Introduction}
Medical image segmentation remains a highly active area of research, evidenced by the U-Net architecture receiving over 20,000 citations in 2023 alone \cite{ronneberger2015u}.
The introduction of nnU-Net in 2018 was a pivotal moment, highlighting that careful implementation and configuration of the architecture are more crucial for achieving state-of-the-art results than  modifying the architecture itself \cite{isensee2018nnu,isensee2021nnu}.
Despite this, the attraction of innovative architectures from the broader computer vision domain, such as Transformers \cite{vaswani2017attention} and Mamba \cite{gu2023mamba}, persists. Adaptations of these cutting-edge designs to the medical imaging domain have emerged, with claims of superior performance over the conventional CNN-based U-Net \cite{swinunetr,chen2021transunet,hatamizadeh2022unetr,he2023swinunetrv2,zhou2021nnformer,xing2024segmamba,ma2024u,cao2022swin,gao2021utnet,xie2021cotr,wang2021transbts}. \\\\
In this paper, we critically examine these claims and find that the current rapid adoption of new methods to the medical domain comes with a lack of stringent validation. As a consequence, we observe that many recent claims of methodological superiority do not hold when systematically tested in a comprehensive benchmark. This trend raises significant concerns, indicating a prevailing attention bias in medical image segmentation towards novel architectures. To overcome this bias and redirect the field towards meaningful methodological progress, we call for a systemic change emphasizing rigorous validation practices. Our study makes the following contributions:
\begin{enumerate}
    \item We systematically identify validation pitfalls in the field and provide recommendations for how to avoid them.
    \item We conduct a large-scale benchmark under a thorough validation protocol to scrutinize the performance of prevalent segmentation methods.
    \item Based on this analysis, we identify key methodological components for medical image segmentation as well as a set of suitable benchmarking datasets.  
    \item We release a series of updated standardized baselines for 3D medical segmentation at \href{https://github.com/MIC-DKFZ/nnUNet}{https://github.com/MIC-DKFZ/nnUNet}. These are based on a residual encoder U-Net within the nnU-Net framework and tailored to accommodate a spectrum of hardware capabilities ("M", "L", "XL"). 
\end{enumerate}

\section{Validation Pitfalls}
\label{sec:pitfalls}
In the following, we present a collection of predominant validation pitfalls in current practice paired with recommendations on how to avoid them. In Section~\ref{sec:results}, we underscore the critical need for this initiative by empirically demonstrating how these pitfalls lead to unsupported claims of methodological superiority.

\subsection{Baseline-related Pitfalls} 
Providing a fair and comprehensive comparison to existing work is essential for scientific progress. Currently, we observe a lack of rigour in ensuring meaningful comparison.
\\\\ 
\textbf{P1: Coupling the claimed innovation with confounding performance boosters:} There are multiple ways to artificially boost a method's performance, obfuscating the real impact of the claimed innovation. One example is coupling the claimed innovation with residual connections in the encoder while the baseline uses a vanilla CNN encoder \cite{ma2024u}. Another example is coupling the claimed innovation with additional training data not used in baselines \cite{unetr}. This is even more critical, if the usage of additional data is not made transparent\cite{swinunetr_data}. A related pitfall is to couple the claimed innovation with self-supervised pretraining, while the baselines train from scratch \cite{swinunetr}. A third example is coupling the claimed innovation with larger hardware capabilities, i.e. comparing against baselines that are not scaled to the same compute budget (VRAM usage and training time) \cite{roy2023mednext}. Finally, sometimes claimed innovations are based solely on leaderboard results where the method is coupled with 20-fold ensembling, while other leaderboard entries do not use such costly performance boosters\cite{swinunetr,unetr}. \textbf{Recommendation (R1):} Meaningful validation entirely isolates the effect of the claimed innovation by ensuring a fair comparison to baselines where the proposed method is not coupled with confounding performance boosters. 
\\\\
\textbf{P2: Lack of well-configured and standardized baselines:} nnU-Net has demonstrated that proper method configuration often impacts performance more significantly than the architecture itself \cite{isensee2021nnu}. This suggests that claims of methodological superiority may be misleading if based on comparisons against an ill-configured baseline (i.e. a manually configured U-Net \cite{ronneberger2015u} with nontransparent and potentially subpar hyperparameter optimization). Some methods, like nnU-Net, address the "faulty baseline" problem by offering automatic, high-quality, and thus \textit{standardized}, configuration on new datasets. Despite this, many studies continue to claim methodological superiority  without benchmarking against any such standardized baseline with a proven high-quality configuration \cite{xing2024segmamba,xie2021cotr,wang2021transbts,gao2021utnet,cao2022swin,chen2021transunet,TransFuse}. Beyond auto-configuration frameworks like nnU-Net, it is almost impossible to ensure a high-quality configuration when including existing methods as baselines, because typically no instructions for adaptation to new tasks are provided. This need for manual adjustments, even if equal hyperparameter tuning budget is allocated to all methods, is an error-prone process that ultimately diminishes the relevance of results. \textbf{Recommendation (R2):} Beyond the call for ensuring high-quality configuration of baselines, long-term standardization in the field can only be achieved if newly proposed methods are equipped with adaptation instructions, or ideally, are carefully integrated within auto-configuration frameworks to inherit their capabilities.
%
%
\subsection{Dataset-related Pitfalls} 
\textbf{P3: Insufficient quantity and suitability of datasets:} The nnU-Net study contains experiments demonstrating 1) the vast diversity of biomedical datasets and 2) the corresponding need of testing on a sufficient number and variety of datasets when making claims about general methodological advancements \cite{isensee2021nnu}. However, the median number of datasets employed in recent studies claiming superior segmentation performance is three \cite{swinunetr,unetr,dints,segresnet,he2023swinunetrv2,ma2024u,xing2024segmamba,huang2023stu,roy2023mednext,zhou2021nnformer,xie2021cotr,gao2021utnet,cao2022swin,chen2021transunet}. Though the number might seem unremarkable on its own, it becomes concerning when considering the varying benchmarking suitability of popular datasets. For instance, as we empirically analyse in Section~\ref{sec:results}, neither of the two datasets BTCV \cite{landman20152015} and BraTS \cite{brats2021_1,brats2021_2,brats2021_3}, while being useful environments for solving their respective clinical task, provide a reliable foundation for assessing general methodological advancements. This is due to a high statistical variance (BTCV) and a low systematic variance (BraTS). Despite this, numerous studies claim methodological superiority while at least $50\%$ of the benchmark is made up by either BTCV \cite{xie2021cotr,chen2021transunet,cao2022swin} or BraTS \cite{hatamizadeh2021swin,wang2021transbts,roy2023mednext,segresnet,unetr}. \textbf{Recommendation (R3):} Meaningful validation requires that utilized datasets are a suitable basis for measuring the claimed methodological advancement. This may include sufficient dataset quantity and diversity, as well as benchmarking suitability of individual datasets, as assessed in our study in Section~\ref{sec:results}.
\\\\
\textbf{P4: Inconsistent reporting practices:} 
Standardization of public leaderboard submissions is limited, e.g. allowing varying strategies for ensembling, test time augmentations and post-processing techniques. While perfectly serving the need to demonstrate that a proposed method can push the state-of-the-art when equipped with all bells and whistles, such non-standardized settings undermine the ability to draw meaningful methodological conclusions. Consequently, researchers often resort to custom train/test splits for controlled comparisons against baselines, but these typically involve small test sets that introduce substantial result instability and question the significance of minor performance gains \cite{wu2023d,xu2023levit,zhou2021nnformer}. Further, the practice of reporting selective results only for specific classes from datasets without justifiable reasons further compromises result integrity \cite{zhou2021nnformer,chen2021transunet,cao2022swin,wu2023d,xu2023levit}. \textbf{Recommendation (R4):} 5-fold cross-validation with a rotating validation set improves reliability and often represents a pragmatic solution. However, using the same dataset(s) for development and validation bears the risk of implicit overfitting and a lack of generalizability. Thus, ideally, differentiating between a pool of development datasets and an independent pool of test datasets for cross-validation against baselines would offer a more reliable assessment of method performance.

\section{Systematic 3D Medical Segmentation Benchmark}
Taking these pitfalls and recommendations into account, we revisit recently proposed methods on basis of a systematic and comprehensive benchmark.

\subsection{Compared Methods}
We categorize methods into CNN-based, Transformer-based, and Mamba-based. 
\textbf{CNN-based:} We include nnU-Net's original configuration using a vanilla U-Net as well as a variant employing a U-Net with residual connections in the encoder ("nnU-Net ResEnc") which has been part of the official repository since 2019 \cite{isensee2019attempt}. In the spirit of avoiding benchmarking of unequal hardware settings in the future (see P1), we introduce new nnU-Net ResEnc presets, which use nnU-Net's existing automatic adaptation of batch and patch sizes to target varying VRAM budgets ("M", "L", "XL"). We further include MedNeXt, a transformer-inspired CNN-modification using ConvNeXt blocks (we test size "L" with kernel sizes "k3" and upkernel "k5") \cite{roy2023mednext}, and STU-Net, a series of scaled up U-Nets with increasing parameter counts named "S"(mall), "B"(ase), "L"(arge), and "H"(uge) \cite{huang2023stu}. 
\textbf{Transformer-based:} We test the SwinUNETR's original version \cite{swinunetr} as well as version 2 \cite{he2023swinunetrv2}, nnFormer \cite{zhou2021nnformer}, and CoTr, a hybrid architecture combining convolutional and transformer modules \cite{xie2021cotr}. 
\textbf{Mamba-based:} We test the recently proposed U-Mamba model \cite{ma2024u} employing Mamba-layers either in the U-Net encoder ("U-Mamba Enc"), or exclusively in the bottleneck ("U-Mamba Bot"). We also include an ablation missing in the original publication using the identical setting while switching off the mamba layers ("No-Mamba Base"). All aforementioned methods were originally implemented in the nnU-Net framework except SwinUNETR(V1+V2), which we integrate into the nnU-Net framework due to incomplete configuration instructions (P2). \textbf{Framework comparison: }In addition to comparing recent methods, we also benchmark nnU-Net against a recent alternative framework: Auto3DSeg(Version $1.3.0$)\cite{auto3Dseg,segresnet,dints,swinunetr} is part of the MONAI eco-system \cite{cardoso2022monai} and recently created a buzz at MICCAI 2023 by winning several highly competitive challenges like KiTS2023, thereby positioning itself as an alternative to nnU-Net promising the same auto-configuration functionality~\cite{auto3Dseg}. The framework is tested by means of three featured architectures ("SegResNet"~\cite{segresnet}, "DiNTS"~\cite{dints}, "SwinUNETR"~\cite{swinunetr}).
\\\\
In the spirit of R1 and R2, we employ a standardized scheme for hyperparameter configuration by either 1) using the self-configuration abilities of methods if available, 2) selecting the configuration closest to the respective dataset if multiple configurations were provided, 3) using the default configuration in case no alternatives were provided or 4) where necessary, decreasing the learning rate until convergence was achieved. All models are trained from scratch. The only exception is SwinUNETR in the Auto3DSeg framework. Altering its default of automatically loading pre-trained weights would have contradicted our hyperparameter configuration scheme. We also employed an equal maximum VRAM budget across all methods by running all trainings on a single NVIDIA A100 with 40GB VRAM. This budget excludes the largest STU-Net variant ("H") from our benchmark.

\subsection{Utilized Datasets}
Our benchmark utilizes six datasets: BTCV \cite{landman20152015}, ACDC \cite{bernard2018deep}, LiTS \cite{bilic2023liver,antonelli2022medical}
BraTS2021 \cite{brats2021_1,brats2021_2,brats2021_3}
KiTS2023 \cite{heller2023kits21}, and AMOS2022 (post challenge Task 2) \cite{ji2022amos}. We selected datasets based on popularity, allowing us to follow R3 and assess the prevalent datasets w.r.t their suitability for method benchmarking. Given that an effective benchmarking dataset should enable measuring consistent signals of methodological differences, we derive two requirements for suitability: 1) low standard deviation (SD) of DSC Scores from the same method across the five folds (intra-method SD) indicating statistical stability and a high signal-to-noise ratio. And 2), high SD across different methods (inter-method SD) indicating meaningful signals of methodological differences, i.e. performance does not saturate too fast on the respective task. Our final suitability score is the ratio of inter-method versus intra-method SD. 
\\\\
Following R4, we report results using 5-fold cross-validation, employing splits generated by nnU-Net and applying these consistently across all methods. Since we do not develop new methods in this study, we refrain from distinguishing between development versus test dataset pools. We report results with the average Dice Similarity Coefficients (DSC) being our primary metric, and the Normalized Surface Dice (NSD) as our secondary metric. For both metrics, results are averaged over all classes of each dataset as well as over the five folds to assess generalist segmentation capabilities without delving into problem-specific metric nuances. For datasets featuring hierarchical evaluation regions (BraTS2021, KiTS2023), we calculate metrics for these regions rather than the non-overlapping classes.
\section{Results and Discussion}
\label{sec:results}
%
%
%
\begin{figure}[t!]
    \centering
    \includegraphics[width=.65\textwidth]{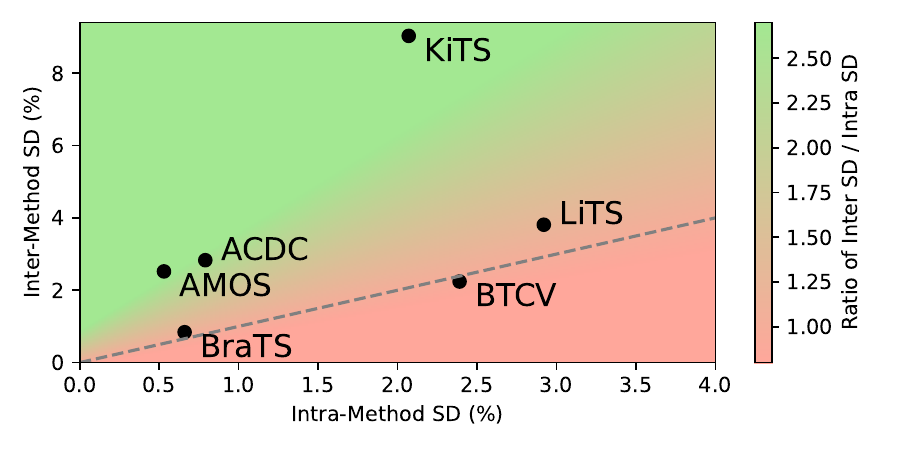}
    \caption{Benchmarking suitability of popular datasets measured as the ratio of inter- versus intra-method standard deviation (SD). The dashed line denotes a ratio of 1.}
    \label{fig:dataset_plot}
\end{figure}
\textbf{KiTS, AMOS, and ACDC are the most suitable datasets for benchmarking 3D segmentation methods.} 
Fig. \ref{fig:dataset_plot} shows the outcome of the dataset analysis based on our benchmark (for detailed results see Appendix Table~\ref{apx:tab:InterDatasetVariation}). We find that KiTS, AMOS, and ACDC exhibit low statistical noise (intra-method SD) while effectively differentiating between methods, as indicated by a high inter-method SD. Out of the three, KiTS features by far the highest inter-method SD, indicating lowest performance saturation on the task. Conversely, scores on BraTS21 are saturated, with minimal variation both between and within methods. BTCV exhibits a SD ratio below one, indicating statistical noise may exceed the signal of performance differences between methods. LiTS represents a middle ground in terms of benchmarking suitability. In summary, ACDC, AMOS, and KiTS can be recommended as the most suitable datasets for benchmarking, BraTS, LiTS, and BTCV are observed to be less suitable for this purpose. \\\\
\begin{table}[t!]
\centering
\caption{Benchmark results of prevalent 3D medical segmentation methods measured as DSC score $[\%]$. Colors are greener for higher scores, normalized per column. Abbreviations, "RT": Runtime measured in GPU hours on A100 40GB PCIe, "Arch.": Architecture, "TF": Transformer, "Mam": Mamba, "A3DS": Auto3DSeg, "AC": Auto-configuration, "nnU": implemented in nnU-Net framework, "DSC": Dice Similarity Coefficient.}
\label{tab:benchmark-results}
\scalebox{0.85}{
\begin{tabular}{lcccccc|cccc}
\toprule
 & BTCV & ACDC & LiTS & BraTS & KiTS & AMOS & VRAM & RT & Arch. & nnU \\
 & n=30 & n=200 & n=131 & n=1251 & n=489 & n=360 & [GB]  & [h] &  &  \\
\midrule
nnU-Net (org.) \cite{isensee2021nnu} & {\cellcolor[HTML]{CEE78C}} \color[HTML]{000000} 83.08 & {\cellcolor[HTML]{B7E88F}} \color[HTML]{000000} 91.54 & {\cellcolor[HTML]{C1E78E}} \color[HTML]{000000} 80.09 & {\cellcolor[HTML]{AFE890}} \color[HTML]{000000} 91.24 & {\cellcolor[HTML]{B0E890}} \color[HTML]{000000} 86.04 & {\cellcolor[HTML]{BCE78F}} \color[HTML]{000000} 88.64 & 7.70 & 9 & CNN & Yes \\
nnU-Net ResEnc M & {\cellcolor[HTML]{C9E78D}} \color[HTML]{000000} 83.31 & {\cellcolor[HTML]{AFE890}} \color[HTML]{000000} 91.99 & {\cellcolor[HTML]{B8E88F}} \color[HTML]{000000} 80.75 & {\cellcolor[HTML]{AFE890}} \color[HTML]{000000} 91.26 & {\cellcolor[HTML]{ACE891}} \color[HTML]{000000} 86.79 & {\cellcolor[HTML]{B9E88F}} \color[HTML]{000000} 88.77 & 9.10 & 12 & CNN & Yes \\
nnU-Net ResEnc L & {\cellcolor[HTML]{C7E78D}} \color[HTML]{000000} 83.35 & {\cellcolor[HTML]{B4E890}} \color[HTML]{000000} 91.69 & {\cellcolor[HTML]{ACE891}} \color[HTML]{000000} 81.60 & {\cellcolor[HTML]{B5E890}} \color[HTML]{000000} 91.13 & {\cellcolor[HTML]{A5E892}} \color[HTML]{000000} 88.17 & {\cellcolor[HTML]{AAE891}} \color[HTML]{000000} 89.41 & 22.70 & 35 & CNN & Yes \\
nnU-Net ResEnc XL & {\cellcolor[HTML]{C9E78D}} \color[HTML]{000000} 83.28 & {\cellcolor[HTML]{B9E88F}} \color[HTML]{000000} 91.48 & {\cellcolor[HTML]{B2E890}} \color[HTML]{000000} 81.19 & {\cellcolor[HTML]{B2E890}} \color[HTML]{000000} 91.18 & {\cellcolor[HTML]{A3E892}} \color[HTML]{000000} \textbf{88.67} & {\cellcolor[HTML]{A4E892}} \color[HTML]{000000} 89.68 & 36.60 & 66 & CNN & Yes \\
\midrule
MedNeXt L k3 \cite{roy2023mednext} & {\cellcolor[HTML]{AAE891}} \color[HTML]{000000} 84.70 & {\cellcolor[HTML]{A3E892}} \color[HTML]{000000} \textbf{92.65} & {\cellcolor[HTML]{A5E892}} \color[HTML]{000000} 82.14 & {\cellcolor[HTML]{AAE891}} \color[HTML]{000000} 91.35 & {\cellcolor[HTML]{A4E892}} \color[HTML]{000000} 88.25 & {\cellcolor[HTML]{A5E892}} \color[HTML]{000000} 89.62 & 17.30 & 68 & CNN & Yes \\
MedNeXt L k5 \cite{roy2023mednext} & {\cellcolor[HTML]{A3E892}} \color[HTML]{000000} \textbf{85.04} & {\cellcolor[HTML]{A3E892}} \color[HTML]{000000} 92.62 & {\cellcolor[HTML]{A3E892}} \color[HTML]{000000} \textbf{82.34} & {\cellcolor[HTML]{A3E892}} \color[HTML]{000000} \textbf{91.50} & {\cellcolor[HTML]{A7E891}} \color[HTML]{000000} 87.74 & {\cellcolor[HTML]{A3E892}} \color[HTML]{000000} \textbf{89.73} & 18.00 & 233 & CNN & Yes \\
\midrule
STU-Net S \cite{huang2023stu} & {\cellcolor[HTML]{D0E78C}} \color[HTML]{000000} 82.92 & {\cellcolor[HTML]{C1E78E}} \color[HTML]{000000} 91.04 & {\cellcolor[HTML]{D6E78B}} \color[HTML]{000000} 78.50 & {\cellcolor[HTML]{D1E78C}} \color[HTML]{000000} 90.55 & {\cellcolor[HTML]{B6E890}} \color[HTML]{000000} 84.93 & {\cellcolor[HTML]{C8E78D}} \color[HTML]{000000} 88.08 & 5.20 & 10 & CNN & Yes \\
STU-Net B \cite{huang2023stu} & {\cellcolor[HTML]{CEE78C}} \color[HTML]{000000} 83.05 & {\cellcolor[HTML]{BCE78F}} \color[HTML]{000000} 91.30 & {\cellcolor[HTML]{CDE78D}} \color[HTML]{000000} 79.19 & {\cellcolor[HTML]{C3E78E}} \color[HTML]{000000} 90.85 & {\cellcolor[HTML]{AFE890}} \color[HTML]{000000} 86.32 & {\cellcolor[HTML]{BFE78E}} \color[HTML]{000000} 88.46 & 8.80 & 15 & CNN & Yes \\
STU-Net L \cite{huang2023stu} & {\cellcolor[HTML]{C7E78D}} \color[HTML]{000000} 83.36 & {\cellcolor[HTML]{BCE78F}} \color[HTML]{000000} 91.31 & {\cellcolor[HTML]{BEE78F}} \color[HTML]{000000} 80.31 & {\cellcolor[HTML]{AFE890}} \color[HTML]{000000} 91.26 & {\cellcolor[HTML]{B1E890}} \color[HTML]{000000} 85.84 & {\cellcolor[HTML]{ACE891}} \color[HTML]{000000} 89.34 & 26.50 & 51 & CNN & Yes \\
\midrule
SwinUNETR \cite{swinunetr} & {\cellcolor[HTML]{FFCA8F}} \color[HTML]{000000} 78.89 & {\cellcolor[HTML]{BCE78F}} \color[HTML]{000000} 91.29 & {\cellcolor[HTML]{F1E688}} \color[HTML]{000000} 76.50 & {\cellcolor[HTML]{CBE78D}} \color[HTML]{000000} 90.68 & {\cellcolor[HTML]{C9E78D}} \color[HTML]{000000} 81.27 & {\cellcolor[HTML]{FFCA8F}} \color[HTML]{000000} 83.81 & 13.10 & 15 & TF & Yes \\
SwinUNETRV2 \cite{he2023swinunetrv2} & {\cellcolor[HTML]{FEE686}} \color[HTML]{000000} 80.85 & {\cellcolor[HTML]{AFE890}} \color[HTML]{000000} 92.01 & {\cellcolor[HTML]{DFE78A}} \color[HTML]{000000} 77.85 & {\cellcolor[HTML]{C8E78D}} \color[HTML]{000000} 90.74 & {\cellcolor[HTML]{BAE78F}} \color[HTML]{000000} 84.14 & {\cellcolor[HTML]{F2E688}} \color[HTML]{000000} 86.24 & 13.40 & 15 & TF & Yes \\
nnFormer \cite{zhou2021nnformer} & {\cellcolor[HTML]{FDE686}} \color[HTML]{000000} 80.86 & {\cellcolor[HTML]{A7E891}} \color[HTML]{000000} 92.40 & {\cellcolor[HTML]{E5E789}} \color[HTML]{000000} 77.40 & {\cellcolor[HTML]{E2E78A}} \color[HTML]{000000} 90.22 & {\cellcolor[HTML]{E5E789}} \color[HTML]{000000} 75.85 & {\cellcolor[HTML]{FFA79B}} \color[HTML]{000000} 81.55 & 5.70 & 8 & TF & Yes \\
CoTr \cite{xie2021cotr} & {\cellcolor[HTML]{E6E789}} \color[HTML]{000000} 81.95 & {\cellcolor[HTML]{C9E78D}} \color[HTML]{000000} 90.56 & {\cellcolor[HTML]{CEE78C}} \color[HTML]{000000} 79.10 & {\cellcolor[HTML]{C9E78D}} \color[HTML]{000000} 90.73 & {\cellcolor[HTML]{B8E88F}} \color[HTML]{000000} 84.59 & {\cellcolor[HTML]{C9E78D}} \color[HTML]{000000} 88.02 & 8.20 & 18 & TF & Yes \\
\midrule
No-Mamba Base & {\cellcolor[HTML]{C0E78E}} \color[HTML]{000000} 83.69 & {\cellcolor[HTML]{B1E890}} \color[HTML]{000000} 91.89 & {\cellcolor[HTML]{BAE78F}} \color[HTML]{000000} 80.57 & {\cellcolor[HTML]{AFE890}} \color[HTML]{000000} 91.26 & {\cellcolor[HTML]{B1E890}} \color[HTML]{000000} 85.98 & {\cellcolor[HTML]{B2E890}} \color[HTML]{000000} 89.04 & 12.0 & 24 & CNN & Yes \\
U-Mamba Bot \cite{ma2024u} & {\cellcolor[HTML]{C4E78E}} \color[HTML]{000000} 83.51 & {\cellcolor[HTML]{B3E890}} \color[HTML]{000000} 91.79 & {\cellcolor[HTML]{BDE78F}} \color[HTML]{000000} 80.40 & {\cellcolor[HTML]{AFE890}} \color[HTML]{000000} 91.26 & {\cellcolor[HTML]{AFE890}} \color[HTML]{000000} 86.22 & {\cellcolor[HTML]{B0E890}} \color[HTML]{000000} 89.13 & 12.40 & 24 & Mam & Yes \\
U-Mamba Enc \cite{ma2024u} & {\cellcolor[HTML]{DCE78B}} \color[HTML]{000000} 82.41 & {\cellcolor[HTML]{BDE78F}} \color[HTML]{000000} 91.22 & {\cellcolor[HTML]{BEE78E}} \color[HTML]{000000} 80.27 & {\cellcolor[HTML]{C0E78E}} \color[HTML]{000000} 90.91 & {\cellcolor[HTML]{AFE890}} \color[HTML]{000000} 86.34 & {\cellcolor[HTML]{C1E78E}} \color[HTML]{000000} 88.38 & 24.90 & 47 & Mam & Yes \\
\midrule
A3DS SegResNet \cite{auto3Dseg,segresnet} & {\cellcolor[HTML]{FFE487}} \color[HTML]{000000} 80.69 & {\cellcolor[HTML]{C7E78D}} \color[HTML]{000000} 90.69 & {\cellcolor[HTML]{CBE78D}} \color[HTML]{000000} 79.28 & {\cellcolor[HTML]{C6E78D}} \color[HTML]{000000} 90.79 & {\cellcolor[HTML]{C9E78D}} \color[HTML]{000000} 81.11 & {\cellcolor[HTML]{DAE78B}} \color[HTML]{000000} 87.27 & 20.00 & 22 & CNN & No \\
A3DS DiNTS \cite{auto3Dseg,dints} & {\cellcolor[HTML]{FFBF93}} \color[HTML]{000000} 78.18 & {\cellcolor[HTML]{FFAA9A}} \color[HTML]{000000} 82.97 & {\cellcolor[HTML]{FFAB9A}} \color[HTML]{000000} 69.05 & {\cellcolor[HTML]{FFA79B}} \color[HTML]{000000} 87.75 & {\cellcolor[HTML]{FFD28D}} \color[HTML]{000000} 65.28 & {\cellcolor[HTML]{FFB397}} \color[HTML]{000000} 82.35 & 29.20 & 16 & CNN & No \\
A3DS SwinUNETR \cite{auto3Dseg,swinunetr} & {\cellcolor[HTML]{FFA79B}} \color[HTML]{000000} 76.54 & {\cellcolor[HTML]{FFA79B}} \color[HTML]{000000} 82.68 & {\cellcolor[HTML]{FFA79B}} \color[HTML]{000000} 68.59 & {\cellcolor[HTML]{F2E688}} \color[HTML]{000000} 89.90 & {\cellcolor[HTML]{FFA79B}} \color[HTML]{000000} 52.82 & {\cellcolor[HTML]{FFDD89}} \color[HTML]{000000} 85.05 & 34.50 & 9 & TF & No \\
\bottomrule
\end{tabular}
}
\end{table}
\textbf{CNN-based U-Nets yield best performance.}
Table~\ref{tab:benchmark-results} shows our experimental results (see Appendix Table~\ref{appendix:NSD_table} for results measured as NSD). CNN-based U-Nets implemented in nnU-Net consistently deliver strong performance across all six datasets. Besides the original nnU-Net, this includes STUNet, ResEnc M/L/XL, MedNeXt and No-Mamba base. MedNeXt consistently stands out with the best performance on all datasets except KiTS, although the gaps are smaller on the datasets with higher benchmarking suitability. Furthermore, MedNeXt's performance gains come at a substantial cost of increased training time (especially k5). Additional experiments in Appendix Table \ref{appendix:dice_score_ablation} indicate that parts of MedNeXt's advantages can be explained by target spacing selection and are thus not exclusively linked to a superior architecture. Given that STU-Net was primarily introduced with a focus on transfer learning, we analysed the effect of pre-training on the Totalsegmentator dataset in Appendix Table \ref{appendix:pretraining_ablation}  \cite{totalseg}. Importantly, the observed superiority of CNNs is tied to the current experimental setting of training methods from scratch on benchmarks of limited size. While the advantages of Transformers observed in other imaging domains have not yet been realized in medical imaging, the future may hold promise for their success as larger scales of training data become available and transfer learning techniques improve.
In contrast to prior claims, Transformer-based architectures (SwinUNETR, nnFormer, CoTr) fail to match the performance of CNNs. This includes not matching performance of the original nnU-Net, which has been released long before the Transformer-based architectures. CoTr shows the best results in the Transformer category, which prior literature related to its convolutional components \cite{roy2023transformer}. 
%
U-Mamba initially appears to perform well across segmentation tasks, but comparison against the previously missing baseline "No-Mamba Base" reveals that the mamba layers actually have no effect on performance, and instead, the originally reported gains were due to coupling the method with a residual U-Net (see P1).
The fact that SegResNet shows best performance among methods implemented in Auto3DSeg underscores that the observed superiority of CNNs is not merely a bias introduced by nnU-Net. 
\\\\
\textbf{nnU-Net is the state-of-the-art segmentation framework.} We find that none of the three methods featured in Auto3DSeg reaches the original nnU-Net baseline ("org.") performance, indicating a substantial disadvantage due to the underlying Auto3DSeg framework. This negative gap occurs despite significantly lower VRAM usage and training time of the nnU-Net baseline. When comparing the two frameworks with an identical method (SwinUNETR), nnU-Net wins in 5 out of 6 datasets. Following an official Auto3DSeg tutorial~\cite{auto3Dseg_kits_tut} we improved the results via manual changes in configuration and further increasing its computing budget, but failed to reach competitive performance (see Appendix Table~\ref{tab:auto3dseg-ablation-results}). Taken together, while Auto3DSeg can be pushed to produce state-of-the-art results, as evidenced by its recent challenge wins, its out-of-the-box capabilities do not match nnU-Net.\\\\
\textbf{Scaling models is important especially on larger datasets}
We tested the effect of model scaling based on two methods: nnU-Net Resenc M/L/XL and STU-Net S/B/L. We find that on the more challenging tasks AMOS and KiTS, a significant boost in performance is observed as the compute budget increases. As expected, the "easier" tasks BTCV and BraTS bear less potential for performance gains from model scaling. These findings underscore the importance of size-awareness and dataset-awareness for meaningful method comparison. For instance, evidence for the superiority of a large new segmentation model should not be based on comparison against a much smaller original nnU-Net.

\section{Conclusion}
\label{sec:discussion}
Our benchmark reveals a concerning trend in 3D medical image segmentation: most methods introduced in recent years fail to surpass the original nnU-Net baseline introduced in 2018. This raises the question: How can we steer the field towards genuine progress? In this study, we link the observed shortcomings to a widespread lack of rigor in method validation. To counteract this, we introduce: 1) A systematic collection of validation pitfalls along with recommendations for their avoidance, 2) The release of updated standardized baselines facilitating meaningful method validation, 3) A strategy for measuring the suitability of datasets for method benchmarking. 
\\\\
Beyond these contributions, achieving true and lasting progress in the field requires a cultural shift, where the quality of validation is valued as much as the novelty of network architectures. Making this shift happen will be the responsibility of method developers, users, and reviewers alike. 
\begin{credits}
\subsubsection{\ackname}
This work was partly funded by Helmholtz Imaging (HI), a platform of the Helmholtz Incubator on Information and Data Science.
\subsubsection{\discintname}
The authors have no competing interests to declare that are
relevant to the content of this article.
\end{credits}

\bibliographystyle{abbrv}
\bibliography{Paper-2847}
\newpage
\appendix

\begin{table}[t!]
\centering
\caption{Ablation of Auto3DSeg SegResNet\cite{auto3Dseg,segresnet} following tutorial~\cite{auto3Dseg_kits_tut} using fold 0 of KiTS. All experiments were executed on A100 GPUs with 40GB of VRAM. Following the authors recommendation to increase compute resources (manual increase of epochs and patch size), A3DS SegResNet yielded significantly improved results, achieving 87.77\% while taking approx. 240 GPU hours (80x30h) and a total of 320GB VRAM (8x40GB). Although this result now surpasses the standard nnU-Net (9h, 7GB VRAM; 86.25\%) it is still outperformed by nnU-Net ResEnc M (11h, 9GB; 87.91\%) and its larger cousins.}
\label{tab:auto3dseg-ablation-results}
\scalebox{0.7}{
\begin{tabular}{ccc|ccccc}
\toprule
Model & GPU hours & VRAM & Epochs & Batch Size & Patch Size & Spacing & KiTS Fold 0 \\
 &  & (GPUs $\times$ MB) &  &  & & &  DSC [\%] \\
\midrule
nnU-Net (org.) & 8.88 & $1 \times 6901$ & - & $2$ & $128\times128\times128$ & $1\times0.78\times0.78$ & 86.25 \\
nnU-Net ResEnc M & 11.39 & $1 \times 8805$ & - & $2$ & $128\times128\times128$ & $1\times0.78\times0.78$ & 87.91 \\
nnU-Net ResEnc L & 35.28 & $1 \times 24223$ & - & $2$ & $160\times224\times192$ & $1\times0.78\times0.78$ & 88.60 \\
\midrule
A3DS SegResNet & 39.72 & $1 \times 20267$ & 300 & $2$ & $144\times224\times224$ & $1\times0.78\times0.78$ & 83.73 \\
A3DS SegResNet & 61.28 & $8\times20267$ & 300 & $8\times2$ & $144\times224\times224$ & $1\times0.78\times0.78$ & 76.81 \\
A3DS SegResNet & 136.64 & $8\times20267$ & 600 & $8\times2$ & $144\times224\times224$ & $0.78\times0.78\times0.78$ & 85.60 \\
A3DS SegResNet & 247.44 & $8\times39873$ & 900  & $8\times2$ & $224\times256\times256$ & $0.78\times0.78\times0.78$ & 87.77 \\
\bottomrule
\end{tabular}
}
\end{table}

\begin{table}
    \centering
    \caption{\textbf{Not all datasets can be recommended to develop and compare architectures.} We report the standard deviation of DSC scores across folds of the same method (\textbf{intra} method SD). We compare this to the standard deviation computed over the average DSC scores across all methods on the dataset (\textbf{inter} method SD). The greater the ratio $\frac{inter_{SD}}{intra_{SD}}$, the more suitable is a dataset for separating methods. "SD": Standard Deviation}
\label{apx:tab:InterDatasetVariation}
\scalebox{.73}{
\begin{tabular}{lcccccc}
\toprule
 & \textbf{BTCV} & \textbf{ACDC} & \textbf{LiTS} & \textbf{BraTS2021} & \textbf{KiTS2023} & \textbf{AMOS2022} \\
\midrule
nnU-Net (org.) & 2.6\% & 0.8\% & 3.5\% & 0.62\% & 2.0\% & 0.43\% \\
nnU-Net ResEnc M & 2.4\% & 0.62\% & 2.6\% & 0.67\% & 2.2\% & 0.57\% \\
nnU-Net ResEnc L & 2.7\% & 0.6\% & 2.4\% & 0.57\% & 1.3\% & 0.59\% \\
nnU-Net ResEnc XL & 2.7\% & 0.51\% & 2.4\% & 0.62\% & 1.2\% & 0.43\% \\
\midrule
MedNeXt L k3 & 2.1\% & 0.26\% & 2.3\% & 0.66\% & 0.94\% & 0.43\% \\
MedNeXt L k5 & 2.0\% & 0.2\% & 2.4\% & 0.59\% & 1.2\% & 0.43\% \\
\midrule
STU-Net S & 2.2\% & 0.6\% & 3.3\% & 0.72\% & 1.7\% & 0.42\% \\
STU-Net B & 2.3\% & 0.78\% & 3.6\% & 1.0\% & 1.9\% & 0.52\% \\
STU-Net L & 2.6\% & 0.85\% & 2.4\% & 0.62\% & 2.1\% & 0.45\% \\
\midrule
SwinUNETR & 2.7\% & 0.65\% & 3.1\% & 0.75\% & 2.0\% & 0.44\% \\
SwinUNETRV2 & 2.1\% & 0.51\% & 2.8\% & 0.55\% & 1.7\% & 0.56\% \\
nnFormer & 2.1\% & 0.21\% & 2.3\% & 0.52\% & 4.2\% & 0.5\% \\
CoTr & 2.8\% & 0.83\% & 2.8\% & 0.69\% & 1.4\% & 0.64\% \\
\midrule
No-Mamba Base & 1.9\% & 0.51\% & 2.9\% & 0.55\% & 2.1\% & 0.32\% \\
U-Mamba Bot & 2.3\% & 0.59\% & 2.1\% & 0.71\% & 2.7\% & 0.43\% \\
U-Mamba Enc & 2.3\% & 0.47\% & 1.7\% & 0.64\% & 2.2\% & 0.5\% \\
\midrule
A3DS SegResNet & 3.0\% & 0.33\% & 2.7\% & 0.52\% & 1.7\% & 0.48\% \\
A3DS DiNTS & 3.0\% & 2.2\% & 2.5\% & 0.79\% & 5.3\% & 1.3\% \\
A3DS SwinUNETR & 1.8\% & 3.6\% & 6.6\% & 0.69\% & 1.5\% & 0.64\% \\
\midrule
\multicolumn{7}{c}{\textbf{Averages}}\\
\midrule
\textbf{Intra} Method SD  & 2.39\% & 0.79\% & 2.89\% & 0.66\% & 2.07\% & 0.53\% \\
\textbf{Inter} Method SD & 2.24\% & 2.83\% & 3.80\% & 0.84\% & 9.03\% & 2.52\% \\
Inter/Intra Ratio & {\cellcolor[HTML]{FFA79B}} \color[HTML]{000000} 94\% & {\cellcolor[HTML]{A4E892}} \color[HTML]{000000} 357\% & {\cellcolor[HTML]{F2E688}} \color[HTML]{000000} 132\% & {\cellcolor[HTML]{F2E688}} \color[HTML]{000000} 127\% & {\cellcolor[HTML]{A4E892}} \color[HTML]{000000} 435\% & {\cellcolor[HTML]{A4E892}} \color[HTML]{000000} 474\% \\
\midrule
\midrule
\multicolumn{7}{c}{\textbf{Averages w/o A3DS}}\\
\midrule
\textbf{Intra} Method SD  & 2.35\% & 0.56\% & 2.66\% & 0.66\% & 1.93\% & 0.48\% \\
\textbf{Inter} Method SD & 1.52\% & 0.57\% & 1.68\% & 0.35\% & 3.14\% & 2.28\% \\
Inter/Intra Ratio & {\cellcolor[HTML]{FFC093}} \color[HTML]{000000} 65\% & {\cellcolor[HTML]{F2E688}} \color[HTML]{000000} 102\% & {\cellcolor[HTML]{FFC093}} \color[HTML]{000000} 63\% & {\cellcolor[HTML]{FFAF98}} \color[HTML]{000000} 53\% & {\cellcolor[HTML]{BCE78F}} \color[HTML]{000000} 163\% & {\cellcolor[HTML]{A3E892}} \color[HTML]{000000} 477\% \\
\midrule
\bottomrule
\end{tabular}
}
\end{table}

\begin{table}[]

    \centering
        \caption{Normalized Surface Dice (NSD) with tolerance 2 mm for all methods and datasets. Reported values are averages over the five-fold cross-validation. NSD was computed using \href{https://github.com/google-deepmind/surface-distance}{https://github.com/google-deepmind/surface-distance}. Relative performance between methods is consistent with the observations based on Dice alone (see Table 1 and Results section).}
    \label{appendix:NSD_table}
    \scalebox{.85}{
    \begin{tabular}{lcccccc}
\toprule
\textbf{Architecture} & \textbf{LiTS} & \textbf{BTCV} & \textbf{ACDC} & \textbf{BraTS2021} & \textbf{KiTS2023} & \textbf{AMOS2022} \\
\midrule
nnU-Net (org.) & {\cellcolor[HTML]{BEE78E}} \color[HTML]{000000} 78.26 & {\cellcolor[HTML]{C6E78D}} \color[HTML]{000000} 85.53 & {\cellcolor[HTML]{B4E890}} \color[HTML]{000000} 94.93 & {\cellcolor[HTML]{B6E88F}} \color[HTML]{000000} 93.64 & {\cellcolor[HTML]{B3E890}} \color[HTML]{000000} 82.91 & {\cellcolor[HTML]{BCE78F}} \color[HTML]{000000} 91.49 \\
nnU-Net ResEnc M & {\cellcolor[HTML]{AFE890}} \color[HTML]{000000} 79.96 & {\cellcolor[HTML]{C0E78E}} \color[HTML]{000000} 86.01 & {\cellcolor[HTML]{ACE891}} \color[HTML]{000000} 95.50 & {\cellcolor[HTML]{B3E890}} \color[HTML]{000000} 93.71 & {\cellcolor[HTML]{AEE891}} \color[HTML]{000000} 84.10 & {\cellcolor[HTML]{B7E88F}} \color[HTML]{000000} 91.72 \\
nnU-Net ResEnc L & {\cellcolor[HTML]{ABE891}} \color[HTML]{000000} 80.39 & {\cellcolor[HTML]{BFE78E}} \color[HTML]{000000} 86.08 & {\cellcolor[HTML]{B1E890}} \color[HTML]{000000} 95.11 & {\cellcolor[HTML]{B9E88F}} \color[HTML]{000000} 93.59 & {\cellcolor[HTML]{A5E892}} \color[HTML]{000000} 85.93 & {\cellcolor[HTML]{ACE891}} \color[HTML]{000000} 92.35 \\
nnU-Net ResEnc XL & {\cellcolor[HTML]{B1E890}} \color[HTML]{000000} 79.64 & {\cellcolor[HTML]{C2E78E}} \color[HTML]{000000} 85.89 & {\cellcolor[HTML]{B4E890}} \color[HTML]{000000} 94.90 & {\cellcolor[HTML]{B8E88F}} \color[HTML]{000000} 93.61 & {\cellcolor[HTML]{A3E892}} \color[HTML]{000000} 86.49 & {\cellcolor[HTML]{A7E892}} \color[HTML]{000000} 92.64 \\
\midrule
MedNeXt L k3 & {\cellcolor[HTML]{A4E892}} \color[HTML]{000000} 81.07 & {\cellcolor[HTML]{A8E891}} \color[HTML]{000000} 87.78 & {\cellcolor[HTML]{A3E892}} \color[HTML]{000000} 96.07 & {\cellcolor[HTML]{ACE891}} \color[HTML]{000000} 93.85 & {\cellcolor[HTML]{A4E892}} \color[HTML]{000000} 86.29 & {\cellcolor[HTML]{A5E892}} \color[HTML]{000000} 92.72 \\
MedNeXt L k5 & {\cellcolor[HTML]{A3E892}} \color[HTML]{000000} 81.26 & {\cellcolor[HTML]{A3E892}} \color[HTML]{000000} 88.18 & {\cellcolor[HTML]{A3E892}} \color[HTML]{000000} 96.09 & {\cellcolor[HTML]{A3E892}} \color[HTML]{000000} 94.04 & {\cellcolor[HTML]{A7E892}} \color[HTML]{000000} 85.67 & {\cellcolor[HTML]{A3E892}} \color[HTML]{000000} 92.86 \\
\midrule
STU-Net S & {\cellcolor[HTML]{D1E78C}} \color[HTML]{000000} 76.20 & {\cellcolor[HTML]{CCE78D}} \color[HTML]{000000} 85.13 & {\cellcolor[HTML]{BEE78F}} \color[HTML]{000000} 94.27 & {\cellcolor[HTML]{CAE78D}} \color[HTML]{000000} 93.26 & {\cellcolor[HTML]{BCE78F}} \color[HTML]{000000} 81.08 & {\cellcolor[HTML]{C8E78D}} \color[HTML]{000000} 90.81 \\
STU-Net B & {\cellcolor[HTML]{C7E78D}} \color[HTML]{000000} 77.33 & {\cellcolor[HTML]{CAE78D}} \color[HTML]{000000} 85.30 & {\cellcolor[HTML]{B9E88F}} \color[HTML]{000000} 94.59 & {\cellcolor[HTML]{BCE78F}} \color[HTML]{000000} 93.54 & {\cellcolor[HTML]{B2E890}} \color[HTML]{000000} 83.08 & {\cellcolor[HTML]{C0E78E}} \color[HTML]{000000} 91.28 \\
STU-Net L & {\cellcolor[HTML]{B9E88F}} \color[HTML]{000000} 78.85 & {\cellcolor[HTML]{C3E78E}} \color[HTML]{000000} 85.81 & {\cellcolor[HTML]{B1E890}} \color[HTML]{000000} 95.12 & {\cellcolor[HTML]{B6E890}} \color[HTML]{000000} 93.66 & {\cellcolor[HTML]{B3E890}} \color[HTML]{000000} 83.02 & {\cellcolor[HTML]{ADE891}} \color[HTML]{000000} 92.30 \\
\midrule
SwinUNETR & {\cellcolor[HTML]{EEE688}} \color[HTML]{000000} 73.06 & {\cellcolor[HTML]{FFD78B}} \color[HTML]{000000} 79.79 & {\cellcolor[HTML]{C0E78E}} \color[HTML]{000000} 94.12 & {\cellcolor[HTML]{CFE78C}} \color[HTML]{000000} 93.16 & {\cellcolor[HTML]{D3E78C}} \color[HTML]{000000} 75.91 & {\cellcolor[HTML]{FFC591}} \color[HTML]{000000} 85.13 \\
SwinUNETRV2 & {\cellcolor[HTML]{D8E78B}} \color[HTML]{000000} 75.38 & {\cellcolor[HTML]{EFE688}} \color[HTML]{000000} 82.52 & {\cellcolor[HTML]{B1E890}} \color[HTML]{000000} 95.15 & {\cellcolor[HTML]{CFE78C}} \color[HTML]{000000} 93.15 & {\cellcolor[HTML]{C0E78E}} \color[HTML]{000000} 80.11 & {\cellcolor[HTML]{F3E688}} \color[HTML]{000000} 88.47 \\
nnFormer & {\cellcolor[HTML]{E0E78A}} \color[HTML]{000000} 74.66 & {\cellcolor[HTML]{F2E688}} \color[HTML]{000000} 82.29 & {\cellcolor[HTML]{A7E892}} \color[HTML]{000000} 95.83 & {\cellcolor[HTML]{CCE78D}} \color[HTML]{000000} 93.22 & {\cellcolor[HTML]{F1E688}} \color[HTML]{000000} 69.43 & {\cellcolor[HTML]{FFA99A}} \color[HTML]{000000} 82.93 \\
CoTr & {\cellcolor[HTML]{C8E78D}} \color[HTML]{000000} 77.25 & {\cellcolor[HTML]{DAE78B}} \color[HTML]{000000} 84.10 & {\cellcolor[HTML]{C6E78D}} \color[HTML]{000000} 93.74 & {\cellcolor[HTML]{BEE78E}} \color[HTML]{000000} 93.49 & {\cellcolor[HTML]{BCE78F}} \color[HTML]{000000} 80.92 & {\cellcolor[HTML]{C9E78D}} \color[HTML]{000000} 90.75 \\
\midrule
No-Mamba Base & {\cellcolor[HTML]{B9E88F}} \color[HTML]{000000} 78.88 & {\cellcolor[HTML]{BEE78E}} \color[HTML]{000000} 86.14 & {\cellcolor[HTML]{AFE890}} \color[HTML]{000000} 95.26 & {\cellcolor[HTML]{B6E88F}} \color[HTML]{000000} 93.64 & {\cellcolor[HTML]{B0E890}} \color[HTML]{000000} 83.56 & {\cellcolor[HTML]{B1E890}} \color[HTML]{000000} 92.08 \\
U-Mamba Bot & {\cellcolor[HTML]{B8E88F}} \color[HTML]{000000} 78.91 & {\cellcolor[HTML]{BBE78F}} \color[HTML]{000000} 86.40 & {\cellcolor[HTML]{ADE891}} \color[HTML]{000000} 95.40 & {\cellcolor[HTML]{B6E88F}} \color[HTML]{000000} 93.65 & {\cellcolor[HTML]{B1E890}} \color[HTML]{000000} 83.27 & {\cellcolor[HTML]{B2E890}} \color[HTML]{000000} 92.00 \\
U-Mamba Enc & {\cellcolor[HTML]{BBE78F}} \color[HTML]{000000} 78.60 & {\cellcolor[HTML]{D3E78C}} \color[HTML]{000000} 84.60 & {\cellcolor[HTML]{BDE78F}} \color[HTML]{000000} 94.33 & {\cellcolor[HTML]{CCE78D}} \color[HTML]{000000} 93.21 & {\cellcolor[HTML]{B0E890}} \color[HTML]{000000} 83.64 & {\cellcolor[HTML]{C0E78E}} \color[HTML]{000000} 91.25 \\
\midrule
A3DS SegResNet & {\cellcolor[HTML]{CFE78C}} \color[HTML]{000000} 76.46 & {\cellcolor[HTML]{F7E687}} \color[HTML]{000000} 82.01 & {\cellcolor[HTML]{C3E78E}} \color[HTML]{000000} 93.88 & {\cellcolor[HTML]{C3E78E}} \color[HTML]{000000} 93.40 & {\cellcolor[HTML]{D5E78C}} \color[HTML]{000000} 75.61 & {\cellcolor[HTML]{DAE78B}} \color[HTML]{000000} 89.85 \\
A3DS DiNTS & {\cellcolor[HTML]{FFAF98}} \color[HTML]{000000} 62.49 & {\cellcolor[HTML]{FFC093}} \color[HTML]{000000} 77.30 & {\cellcolor[HTML]{FFA79B}} \color[HTML]{000000} 83.67 & {\cellcolor[HTML]{FFA79B}} \color[HTML]{000000} 90.36 & {\cellcolor[HTML]{FFCE8E}} \color[HTML]{000000} 58.74 & {\cellcolor[HTML]{FFA79B}} \color[HTML]{000000} 82.75 \\
A3DS SwinUNETR  & {\cellcolor[HTML]{FFA79B}} \color[HTML]{000000} 61.16 & {\cellcolor[HTML]{FFA79B}} \color[HTML]{000000} 74.59 & {\cellcolor[HTML]{FFA99A}} \color[HTML]{000000} 83.94 & {\cellcolor[HTML]{FFDF88}} \color[HTML]{000000} 92.00 & {\cellcolor[HTML]{FFA79B}} \color[HTML]{000000} 46.37 & {\cellcolor[HTML]{FFDB8A}} \color[HTML]{000000} 86.93 \\
\bottomrule
\end{tabular}}
\end{table}

\begin{table}[]
    \centering
    \caption{Ablation of average DSC when using isotropic spacing for the nnU-Net ResEnc L on ACDC and BTCV instead of the default anisotropic spacing. Results indicate that MedNeXt’s performance can in part be explained by its target spacing selection and is thus not exclusively linked to the better architecture. "DSC": Dice similarity coefficient.}
        \label{appendix:dice_score_ablation}
\begin{tabular}{llccc|r}
\toprule
\textbf{Dataset} & \textbf{Method} & \textbf{Patch Size }&\textbf{Spacing} &  \textbf{Batch Size} &    \textbf{DSC} \\
\midrule
\multirow{3}{*}{\textbf{BTCV}} & 
      nnU-Net ResEnc L &   80x256x256 &  3x0.76x0.76 &           2 &  83.35 \\
     & nnU-Net ResEnc L (iso) &  192x192x192 &  1x1x1 &           2 &  84.01 \\
     & MedNeXt L k3 &  128x128x128 &        1x1x1 &           2 &  84.70 \\\midrule
\multirow{3}{*}{\textbf{ACDC}} & 
     nnU-Net ResEnc L &   20x256x224 &  5x1.56x1.56 &          10 &  91.69 \\
     & nnU-Net ResEnc L (iso) &   96x256x256 &  1x1x1 &           3 &  92.64 \\
     & MedNeXt L k3 &  128x128x128 &        1x1x1 &           2 &  92.65 \\\midrule
\multirow{3}{*}{\textbf{AMOS}} & 
     nnU-Net ResEnc L &   96x224x224 &  2x0.71x0.71 &          2 &  89.40 \\
     & nnU-Net ResEnc L (iso) &   192x192x192 &  1x1x1 &           2 &  89.60 \\
     & MedNeXt L k3 &  128x128x128 &        1x1x1 &           2 &  89.62 \\
\bottomrule
\end{tabular}

\end{table}


\begin{table}
\centering
\label{appendix:pretraining_ablation}
\caption{Since STU-Net was presented as a model for transfer learning, we fine-tuned a STU-Net L network, that was pre-trained on the totalsegmentator dataset \cite{totalseg} for 4000 epochs, on the other datasets, analogous to the corresponding publication \cite{huang2023stu}. Fine-tuning on BraTS did not converge using the default fine-tuning learning rate of 0.001.}
\label{tab:STU-Net}
\scalebox{0.85}{
\begin{tabular}{lcccccc|cccc}
\toprule
 & BTCV & ACDC & LiTS & BraTS & KiTS & AMOS & VRAM & RT & Arch. & nnU \\
 & n=30 & n=200 & n=131 & n=1251 & n=489 & n=360 & [GB]  & [h] &  &  \\
\midrule
STU-Net L \cite{huang2023stu} &  83.36 & 91.31 & 80.31 & 91.26 & 85.84 &  89.34 & 26.50 & 51 & CNN & Yes \\
\midrule
STU-Net L pretrained \cite{huang2023stu} &  84.28 & 91.53 & 81.57 & 0& 88.32&  89.46 & 26.50 & 51* & CNN & Yes \\
\bottomrule
\multicolumn{11}{c}{*Fine-tuning runtime only. Pre-training takes about 4 times longer.}\\
\end{tabular}
}
\end{table}

\end{document}